\documentclass[a4paper]{article}

\usepackage[english]{babel}
\usepackage[utf8x]{inputenc}
\usepackage[T1]{fontenc}

\usepackage[a4paper,top=3cm,bottom=2cm,left=3cm,right=3cm,marginparwidth=1.75cm]{geometry}

\usepackage{amsmath}
\usepackage{graphicx}
\usepackage[colorinlistoftodos]{todonotes}
\usepackage[colorlinks=true, allcolors=blue]{hyperref}
\usepackage{relsize}
\usepackage{url}

\def\silnet{\emph{SiluCalNet}}
\def\accnet{\emph{AccuCalNet}}
\def\fullnet{\emph{CaloriNet}}
\def\etal{\emph{et al.}}

\title{\fullnet: From silhouettes to calorie estimation in private environments}
\author{
  Alessandro Masullo\footnote{a.masullo@bristol.ac.uk}
  \and
  Tilo Burghardt
  \and
  Dima Damen
  \and
  Sion Hannuna
  \and
  Victor Ponce-López
  \and
  Majid Mirmehdi
}

\begin{document}
\maketitle

\begin{abstract}
{We propose} a novel {deep fusion architecture,} \fullnet{}, for the {online} estimation of energy expenditure for free living monitoring in {private}  environments, {where} RGB data is {discarded} and replaced by silhouettes. 
        {Our fused convolutional neural network architecture is trainable end-to-end, to estimate calorie expenditure, using temporal foreground silhouettes alongside accelerometer data.}
        %{Our approach} optimally uses temporal silhouettes utilizes deep learning to estimate the {calorie} expenditure using an end-to-end convolutional neural network. The framework we propose allows the \ale{combination of image silhouettes and accelerometers. 
        The network is trained and cross-validated on a publicly available dataset, %~\cite{Tao2017}.
{\em SPHERE\_RGBD + Inertial\_calorie}. % ~\cite{Tao2017}.
        %after the conversion of RGB images into silhouettes. 
        Results show state-of-the-art minimum error on the estimation of energy expenditure {(calories per minute)}, {outperforming} alternative, standard {and single-modal} 
        %deep-learning available 
        techniques. 
\end{abstract}

%-------------------------------------------------------------------------
\section{Introduction}
\label{sec:intro}
Physical activity has been linked to general health \cite{Marshall2011} and has shown positive psychological benefits \cite{Byrne1993} in clinical tests. Further, sedentary behaviour has consequences that may impose many health risks, for example on musculoskeletal health. This is especially important for older adults, for whom physical activity can counteract the detrimental effect on the cardiovascular system and skeletal muscles associated with age \cite{Chodzko-Zajko2009}. {Monitoring the extent of physical activity via} energy expenditure (EE) is therefore of valuable importance and different approaches have been proposed in the literature, from the use of questionnaires \cite{Craig2003}, to metabolic lookup tables {(METs)} \cite{AINSWORTH1993}, to peak oxygen uptake estimations \cite{Armstrong1990}.

With the development of novel technologies, Internet of Things (IoT) is playing an important role in monitoring well being and health \cite{Ray2014}. Accelerometers\footnote{The terms \textit{accelerometers}, \textit{inertial} and \textit{wearable} sensors are used indiscriminately throughout this paper.} have often been adopted for the estimation of EE \cite{Yang2010}, although video monitoring systems have recently showed superior performances \cite{Tao2017}, especially when combined with inertial based measurements \cite{Tao2018}. However, recent works, such as from Birchley \etal \cite{Birchley2017}, Ziefle \etal \cite{Ziefle2011} and Jancke \etal \cite{Jancke2001} have highlighted the important aspect of privacy concern in medical technologies for smart homes, showing a critical view of such systems from participants. Patients often fear misuse of their video recordings, data leakage or loss due to technical issues. These concerns have been addressed in the work from Hall \etal \cite{Hall2016} by replacing the RGB video stream with bounding boxes, skeletons and silhouettes, which not only assess the privacy issue, but also allow to scale the amount of data recorded to a size which is more suitable for an IoT platform.

In this paper, we present a fused convolutional architecture, named \fullnet{}, for the {online} estimation of EE in private environments, where RGB images are discarded after the generation of silhouettes. Our method uses a data-fusion approach by extracting features from image silhouettes and accelerometer data using a convolutional neural net (CNN), and combining them using fully connected layers to estimate the calorie expenditure. {Our approach is based on the evaluation of buffers of data collected over a variable interval of time, allowing an online estimation of calories, rendering the method suitable for energy expenditure monitoring applications.} The method was trained and cross-validated on a publicly available dataset~\cite{Tao2017}. Our results are compared against the latest and most accurate accelerometer EE techniques and more traditional METs lookup tables, obtaining state-of-the-art results.

To stress the importance of our data-fusion approach, we also study the contribution of each modality when used exclusively, by assessing the sub-architectures or branches of our \fullnet{}. We name these branches  \silnet{} and \accnet{}, respectively for the video and the accelerometer modalities alone. While the fusion approach allows a reduction of the overall error from the previous state-of-the-art of 1.21 to 0.88 calories/min, these two modalities are independently able to achieve comparable performances with overall error of 0.98 for \accnet{} and 0.95 for \silnet{}. These sub-architectures are available as standalone alternatives to the fusion approach, making our framework suitable for a vision only or wearable only solution.
%\DimaN{[do you want to add a sentence to say that Silhouette alone achieves 90\% (or whatever the number is) accuracy when compared to fusion data? That would strengthen the intro]} \ale{\textit{Added.}} \mmn{I don't think what you have added Alessandro is quite exactly what Dima was suggesting? Let's talk about this on Thursday. Also, why do we still have these subnets?}

% To further stress the importance of our work, we also apply our method to estimate the EE of a completely free-living healthcare monitoring environment from the data collected by the IRC-SPHERE project \cite{Zhu2015a}. The data consists of silhouettes and accelerometer measurements recorded in one of the participants' house. \mmn{we might save that last bit of experimentation for another paper - but leave this in for now.}

\section{{Background and} Related Works}
\label{sec:related}
The estimation of EE is a very complex problem, as it is not only related to the physical movement of the subject, but also their metabolism, level of fitness, physiology and environmental conditions, e.g. temperature, humidity and barometric pressure \cite{Celi2010}. Considerable effort has been invested in the past for characterizing EE using different types of data, including biometric data (i.e. heart rate monitoring), accelerometers, shoe sensors and cameras. In spite of this variability, EE is strongly correlated with the type of activity which is performed. In 1993, the Compendium of Physical Activities \cite{AINSWORTH1993} presented a table with different physical activities connected to EE, described as the ratio of {working} to resting metabolic rates, i.e.  METs. These data include detailed description of activities with {their corresponding EE values}. While METs tables allow a very quick estimation of EE, the approach is based on averages and is only reliable in a statistical sense. Precise measurements of EE are very individual {dependent}, as different subjects perform activities in distinctive ways and therefore consume a different amount of energy.

To allow an individual-{dependent} measurement, the work from Ceesay \etal \cite{Ceesay1989} proposed a heart rate monitoring method that models their EE. A large body of research has focussed instead on the application of accelerometer data to estimate EE. Some works, such as \cite{Albinali2010} and \cite{Altini2012a}, make use of activity-{dependent} models to predict the EE of patients based on the knowledge of the activity they are performing. For a complete review of accelerometer based EE estimation, the reader {is referred to Altini \etal \cite{Altini2015a}, which investigates the methodologies, sensor numbers and locations to obtain the best EE model. Their work} concludes that one single accelerometer close to the subject's centre of mass, combined with an activity-specific estimation model allows for the most accurate and unobtrusive accelerometer-based EE estimation.

One of the most important steps in the use of accelerometer data is the selection of the features. The accelerometer signals are split into contiguous windows, for which a number of frequency and time domain statistics are evaluated, including average, standard deviation, max/min and correlation coefficients, among others \cite{Ellis2014}. The selection of {such hand-crafted} features allows the application of standard machine learning algorithms like artificial neural networks \cite{Staudenmayer2009a}, random forests \cite{Ellis2014} and other regression models \cite{Pande2015} - with performances strongly dependent on those selected features. {Zhu \etal \cite{Zhu2015c} proposed the application of CNNs where the raw accelerometer signal was directly fed into a CNN which automatically learned the features that then allowed a multilayer perceptron to produce EE estimates with errors} up to 35\% lower than methods previous to it. For this reason, Zhu \etal \cite{Zhu2015c}'s method was selected as the baseline for comparison with our results.

Computer vision {has also been deployed to improve} digital health monitoring systems. For example,  \cite{Kong2012} and \cite{Myers2015} attempted to estimate the calories in food by taking single images or short videos of them, although they needed to interact with the user to allow continuous monitoring. % This is actually calories eaten, not calories burnt, maybe it is confusing? Although it's a nice CV calories referece...
Closer to the topic of this paper, Tao \etal \cite{Tao2018} proposed a vision-based system which estimated calorie expenditure using features extracted from RGB-D image data of humans in action. {They showed} that RGB-D data can be successfully adopted to estimate EE instead of accelerometers. {This work was later extended by replacing their hand-crafted features with CNN-generated features \cite{Wang2018}, showing an overall reduction of the error.}
%, although it produced even better results when combined with {them}. 
However, as already addressed {earlier}, {it may be critical for} healthcare and ambient assisted living (AAL) systems to respect privacy conditions and only provide video sequences in the form of silhouettes \cite{Woznowski2015}. Under such conditions, methods such as  \cite{Tao2018} are not  suitable as they require full RGB-D data to estimate EE.

CNN regression has been successfully applied in computer vision, for example for 3D pose \cite{Mahendran2017}, age estimation \cite{Niu2016} and viewpoint evaluation \cite{Massa2016}. For medical data, CNNs were applied for the segmentation of the cardiac left ventricle, parametrised in terms of location and radius \cite{Tan2016}. 
%In the work from Xie \etal \cite{Xie2016}, a deep-learning regression approach was designed for counting and detecting cells in microscopy images. A fully convolutional regression network was adopted to estimate the density map of the cells and their locations. 
More recently, a general framework for the analysis of medical images was proposed by Gibson \etal \cite{Gibson2018}, to provide a pipeline that allows segmentation, regression {(i.e. prediction of attenuation maps in brain scans)} and image generation using deep learning.

In this paper, we propose a fused deep architecture which enables the {online} estimation of EE in privacy-sensitive settings. 
%The method is described in Section~\ref{sec:proposed_method}, including the {estimation} of {temporal} silhouettes (Section~\ref{sec:gen_sil}), 
%the preprocessing of data (Section~\ref{sec:sil_preprocess}) 
%and the network architecture (Section~\ref{sec:network_architecture}). Section~\ref{sec:data_aug} details the data augmentation while the dataset is described in Section~\ref{sec:dataset}. Finally, the discussion of results can be found in Section~\ref{sec:results} and our implementation details in Section~\ref{sec:imp_det}. 
{The method is described in Section~\ref{sec:proposed_method}, including the {estimation} of {temporal} silhouettes, the network architecture, and the data augmentation. The dataset, our implementation, and our results are presented in Section~\ref{sec:exps}.}

\section{Proposed method}
\label{sec:proposed_method}
We propose \fullnet{} for online EE estimation, based on the fusion of image silhouettes and accelerometers. 
{The proposal builds on the strengths of two modalities for calorie estimation; (1) visual input that can better recognise the action undertaken~\cite{Tao2016}, yet is at times occluded and associated with privacy concerns, and (2) wearable accelerometers that are light to carry and increasingly popular for healthcare monitoring, but require subject cooperation in wearing and charging the sensors. Thus, we propose an architecture that fuses both modalities, and \textit{importantly} only uses the silhouette (i.e. foreground segmentation) from the visual input, as this provides improved privacy for monitoring in private environments \cite{Hall2016}.} 
%\DimaN{[two things: you want to say each modaility is good, not both have shortcomings. Also, find some SPHERE reference or other references that say silhouettes are good for privacy and reference here.]} \ale{\textit{We already stressed the importance of silhouettes for the privacy issue in the introduction, so I simply repeated the same reference}}
%\ale{While silhouettes and accelerations can be independently used for the calories estimation, they both suffer of some disadvantages when used alone. For example, the silhouettes are unable to characterise motion when this is occluded (i.e. subject not facing the camera), while accelerometers do not suffer from this problem. On the other hand, accelerometers are unable to discern complex postures if the subject is not moving, while the silhouettes are more suitable for this task. These problems can affect the EE estimation during specific activities, creating unwanted biases. With the aim tackling these disadvantages, our proposed network fuses both silhouettes and accelerometers, in order to take full advantage of the data available in privacy sensitive environments. The proposed algorithm takes a buffer of both types of data as input and produces an estimation of the calories burnt (cal/per minute) roughly every 30 seconds.}

\subsection{{Temporal Silhouettes for Calorie Estimation}} %Generation of silhouettes}
\label{sec:gen_sil}

%\DimaN{[this shouldn't look like it is designed to work with the SPHERE data solely. That's dangerous. Make sure you rewrite this throughout]}

\begin{figure}
	% C:\Users\am14795\Dropbox\documenti\SPHERE\Papers\BMVC2018\CaloriesCNN\images\figure_silhouette.py
	\begin{center}
		\includegraphics[width=0.5\textwidth]{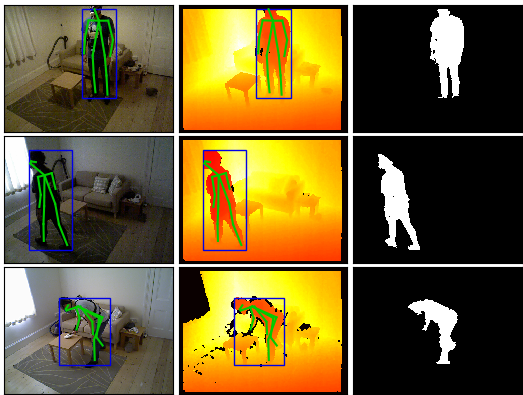}
	\end{center}
	\caption{Examples of silhouettes - % generated from the {\em SPHERE-RGBD + Inertial\_calorie} dataset. 
    Colour and depth images were only used to generate silhouettes and discarded after the process.}
\label{fig:silhouette}
\end{figure}

{To support private environments, we propose to limit the visual input to foreground silhouettes. The method we propose here could use silhouettes extracted from RGB foreground segmentation, or depth-based segmentation as used in our experiments.}
%The {\em SPHERE\_RGBD + Inertial\_calorie} dataset, described in Section~\ref{sec:dataset}, contains RGB-D data, {accelerometers' data {(mounted on the wrist and the waist)}, and groundtruth energy expenditure from a Calorimeter}. The dataset also includes bounding boxes positions surrounding the participants, that we exploited to produce the silhouettes. 
We process the RGB images using OpenPose \cite{Cao2016} to {detect people and }extract the skeletons of the subjects and then perform clustering on the RGB-D values within {each detected bounding box}. 
%Once the clusters were generated, we automatically picked the union of the clusters that were intersecting with the OpenPose skeletons. 
Some generated silhouettes can be found in Figure~\ref{fig:silhouette}. The reader is reminded that RGB-D values are only used to generate the silhouettes and are discarded after this process.

% \subsection{Silhouettes preprocessing}
% \label{sec:sil_preprocess}
% \ale{\textit{This entire section was updated}}

The estimation of EE has a strong dependency on {monitoring duration}, and in particular on the past activities performed. In order to take this into account, {temporal modelling and} dependency must be included in the network architecture. A typical approach for this problem is to feed a large buffer of images into the network as input, but this would demand a large amount of memory. Since the silhouettes only contain binary information, we decided to pursue a different approach and built an average silhouette using a variable number of images. {The idea of transforming a video sequence into a compact representation {(to aid our analysis with CNNs)} is not new, and previous examples of similar propositions can be found in works such as \cite{Bobick2001} and \cite{Bilen2016}}.

{As calorie estimation can be better predicted at various temporal scales, we propose to use a multi-scale temporal template for $N$ time intervals} $\Delta t_N$ of {decreasing} length, so that:
\begin{equation}
\Delta t_1 > \Delta t_2 > ... > \Delta t_N .   
\end{equation}
For each $\Delta t_k$, the silhouettes in the interval $\left[t-\Delta t_k,\ t\right]$ were selected and averaged:
\begin{equation}
\bar{S}_k = \frac{1}{\Delta t_k} {\mathlarger{\sum}_{i=t-\Delta t_k}^{t}} S\left(i\right) . 
\end{equation}
This process produces $N$ {multi-scale temporal} silhouettes $\bar{S}_k$ (one for each $\Delta t$), which were then stacked in a 3D tensor $S^*$, where the $3^{rd}$ dimension is the stacked multi-scale temporal silhouette: 
\begin{equation}
\label{eq:avg_silhouette}
S^*_{t} \equiv \left\lbrace\bar{S}_{1},\ \bar{S}_{2},\ ...,\ \bar{S}_{N}\right\rbrace ~.
\end{equation}
%The {multi-scale temporal} silhouette 
$S^*_t$ is then used for the estimation of the calories at time $t$. This operation allows us to reduce any dependency of the network {on} the choice of the $\Delta t$, facilitating the learning process to pick the correct channels for the best EE estimation {for the various daily actions}.

\subsection{Network architecture}
\label{sec:network_architecture}

\begin{figure}
	\begin{center}
		\includegraphics[width=.82\textwidth]{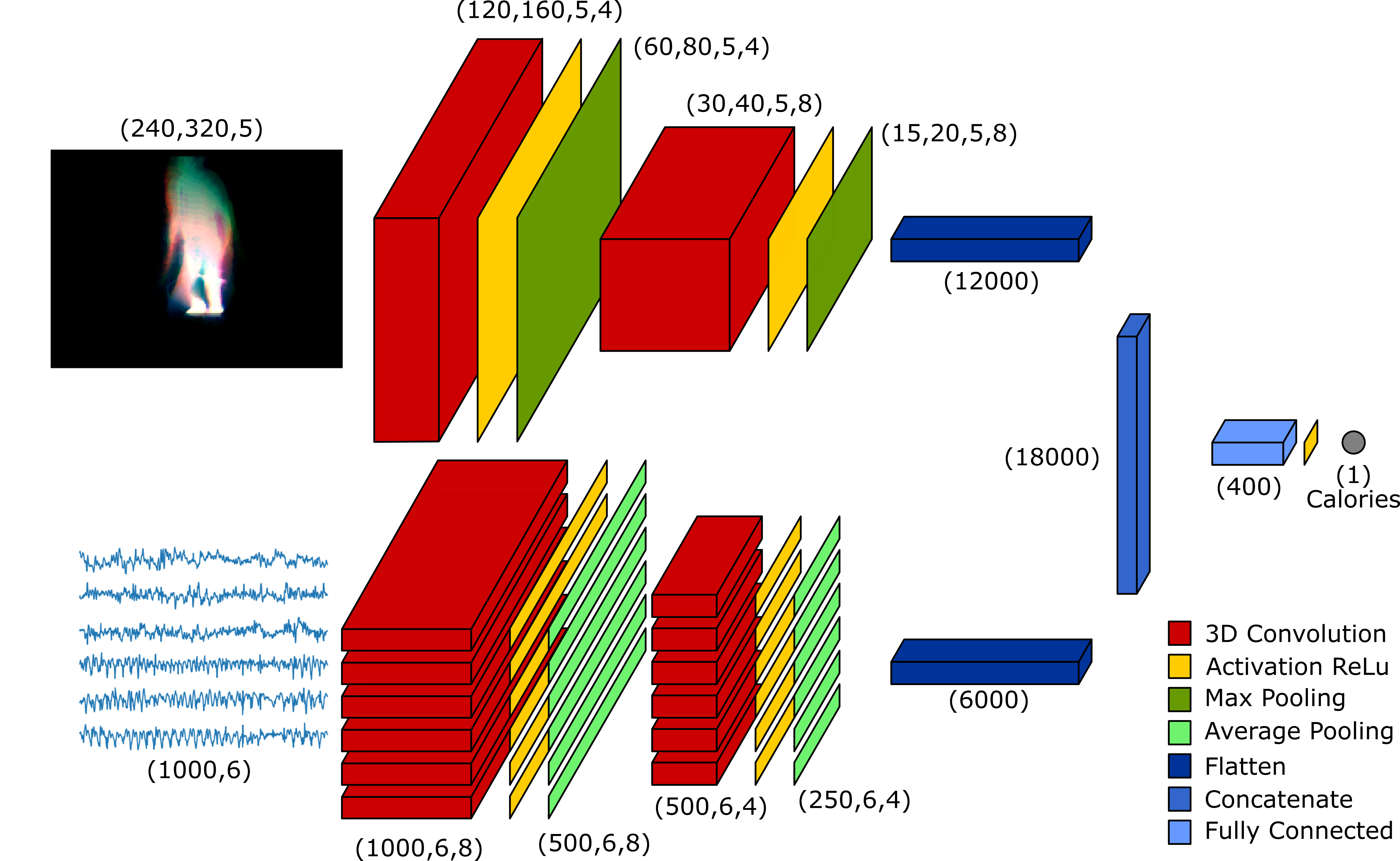}
	\end{center}
	\caption{\fullnet{} -  our architecture combines silhouette data (upper branch, \silnet{}) and accelerometer data (lower branch, \accnet{}) to produce calorie estimation.}
\label{fig:architecture_combined}
\end{figure}

The \fullnet{} architecture is composed of two branches, one for the silhouette data and one for the accelerometer data, as depicted in Figure~\ref{fig:architecture_combined}. The network uses two distinct inputs at time $t$ to produce the calorie estimation $C_t$: the multi-channel average silhouette $S^*_t$ from Eq.~(\ref{eq:avg_silhouette}) and a buffer of accelerometer data in the same time interval $\left[t-\text{max}_k(\Delta t_k),\ t\right]$.

A shallow architecture composed of two stacks of layers was adopted. The features extracted from the silhouettes and acceleration were concatenated and fed into {one fully connected layer} that performs a regression over the calories output. The accelerometer branch was inspired by the work from Zhu \etal \cite{Zhu2015c}, although several modifications were performed to achieve better performances (see Section~\ref{sec:exps} for the implementation details). The silhouettes branch also uses two stacks of layers only. In fact, due to the simplistic nature of the data, being originated from binary foreground images and 6-dimensional accelerometer data, any deeper architecture {is likely to overfit the input}.
%would only result in longer computational time and higher memory requirements. 
We empirically found this depth to suffice for the task of the EE estimation.

{The network is trained end-to-end using} the squared error loss function between the estimated calories $C_p$ and the ground truth $C_{GT}$ {over all times $t$}:
\begin{equation}
\text{Loss}=\sum_t \left(C_p^t-C_{GT}^t\right)^2
\end{equation}

%\DimaN{[it is  not clear whether you accumulate all times individually or per sequence in the training loss]} \ale{\textit{Each multi-channel average silhouette, with a buffer of accelerometer data, is an input datapoint with a single calorie estimation in output. The loss is evaluated between that calorie output and the ground truth. I don't know if keras evaluates the average loss over the batch or over the single training datapoint}}

% \subsection{Investigation of calories only}
% \begin{figure}
% 	\begin{center}
% 		\includegraphics[width=\textwidth]{network_new.png}
% 	\end{center}
% 	\caption{\silnet{}\ architecture for the estimation of calories using silhouettes only.}
% \label{fig:architecture_sil}
% \end{figure}

% \subsection{Network architecture for combined data (\fullnet{})} \label{sec:network_accsil}
% \mmn{Add a  rationale for doing this. Firstly, because we want to see if the added information (which may be very plausible in a healthcare application) does impact the results, and secondly, why it makes sense scientifically, if you can think of a reason!} \ale{\textit{I added a new introduction to this section}}\\

\subsection{Data augmentation}
\label{sec:data_aug}

{Due to the limited training data, as well as to remove any bias in the recording location, we applied the following data augmentation techniques.}

%\subsubsection{Silhouettes}
\noindent \textbf{Silhouettes:} The typical approach for dealing with subjects moving in a frame is to crop the active area and resize it to a fixed size to use as input for the network \cite{He2014}. However, {this is not suitable for temporal silhouettes as the size of the averaged image depends on the motion of the person during the buffered time.}
%due to the large size of the buffer adopted (up to 30 seconds), the subjects' movement often spanned from small regions to the entire size of the frame, therefore we preferred feeding the entire average silhouette image into the network. 
To avoid learning specific positions where actions were performed, data augmentation was implemented. During training, images were randomly flipped (horizontally), tilted, and {translated} (horizontally/vertically). %The transformation matrices producing a rotation $\theta$ and shift of $(t_x,t_y)$ can be represented as:

% \begin{equation}
% T_\theta = \left[\begin{array}{ccc}
%     	\cos(\theta) & -\sin(\theta) & 0 \\ 
%         \sin(\theta) & \cos(\theta) & 0 \\
%     	0			& 0			 & 1 \\
%     \end{array}\right],\ T_\text{shift} = \left[\begin{array}{ccc}
%     	1 & 0 & t_x \\ 
%         0 & 1 & t_y \\
%     	0 & 0 & 1 \\
%     \end{array}\right]
% \end{equation}

The data augmentation parameters adopted were determined empirically (see next section). %and can be found in Section~\ref{sec:imp_det}. 
Although the augmented data sometimes resulted in subjects being {cropped, this matched situations when subjects were only partially in view of the camera}.% or out of frame, the outcome still resembles real-life scenarios  of free-living environments, where people will not necessarily be conscious of the camera and whether they are fully within its view.

%\subsubsection{Accelerometers}
\noindent \textbf{Accelerometers:} For the accelerometer sensors, inspired by the work from Um \etal  \cite{Um2017}, we randomly changed the magnitude of the sensors by multiplying it with a scalar drawn from a Gaussian distribution with mean $1$ and standard deviation $0.1$. In addition, the x-y-z channels of each accelerometer were swapped with random permutations.

\begin{figure} 
	\begin{center}
		\includegraphics[width=\textwidth]{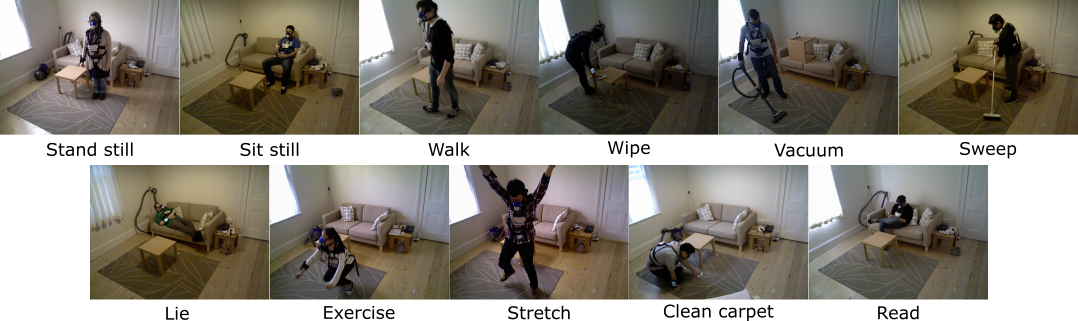}
	\end{center}
	\caption{Sample frames from different subjects and for various activities in the dataset.}
	\label{fig:dataset_activities}
\end{figure}

\section{Experiment Details} \label{sec:exps}

%\subsection{Dataset}
%\label{sec:dataset}
{\bf Dataset ---} {We evaluate our method on the publicly available dataset from~\cite{Tao2017}, {namely {\em SPHERE\_RGBD + Inertial\_calorie}}. This is the only dataset to include RGB-D {and accelerometer input with ground truth calorie measurements obtained from a clinical Calorimeter} for daily activities.
The dataset includes 10 participants, 7 males and 3 females aged between $27.2 \pm 3.8$ years, with average weight of $72.3 \pm 15.0$ kg and average height of $173.6 \pm 9.8$ cm, resulting in average BMI of $23.7 \pm 2.8$. Each participant was recorded with an RGB-D sensor, two accelerometers (mounted on the waist and the arm) and a COSMED K4b2 portable
metabolic measurement system (i.e. a Calorimeter). Eleven activities, as shown in Figure~\ref{fig:dataset_activities}, were performed in a predefined sequence: stand still, sit still, walking, wiping the table, vacuuming, sweeping floor, lying down, exercising, upper body stretching, cleaning stain, reading. 
%For more information about the dataset, please refer to \cite{Tao2018}. 
The dataset presents gaps for some recorded sequences 
%in either the RGB data or the bounding boxes, 
for which we could not generate any silhouettes. Missing data in the training set was therefore replaced by randomly sampling input with the same label from the sequences of the same individual. 

\begin{figure} 
	% C:\Users\am14795\Dropbox\documenti\SPHERE\keras_play\calories\plot_dataset.py
	\begin{center}
		\includegraphics[width=.9\textwidth]{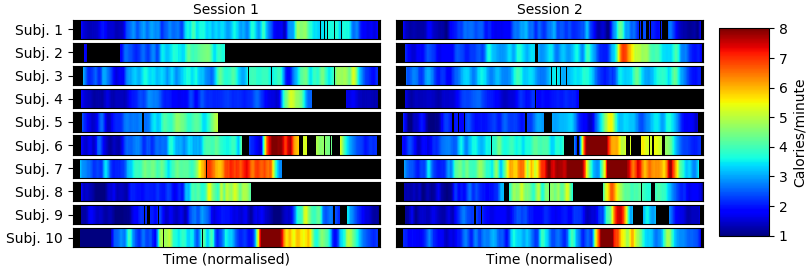}
	\end{center}
	\caption{{Our} visual depiction of the {\small{\em SPHERE\_RGBD + Inertial\_calorie}} dataset. The colour represents the amount of calories/minute, with black areas indicating missing data.}
	\label{fig:dataset}
\end{figure}

Figure~\ref{fig:dataset} presents a visual depiction of the calories recorded in the dataset. Each horizontal {bar} corresponds to one subject performing the same set of activities. Note that {while the calorie measurements} present a certain degree of correlation with the activity performed, each subject has a different response in terms of EE when performing {the same} activity. This difference shows the complexity of the EE problem and highlights the strong limitations of lookup tables when attempting the predict EE for a specific individual.

%\subsection{Implementation details}
%\label{sec:imp_det}
\noindent {\bf Implementation details ---} The network was implemented and trained in Keras using Tensorflow as backend\footnote{{Link to code will be available with the publication}.}. 
%% This next bit is now redundant as the branches were defined earlier.
%In addition to the proposed architecture, \fullnet{}, we also train and assess the two independent branches that use silhouettes only, which we refer to as \silnet{}, and accelerometers only, which we refer to as \accnet{}.
% These networks were built by isolating the two branches in Figure~\ref{fig:architecture_combined} and by feeding them with one single input only.}

\noindent \textbf{\em Silhouettes:}
{The input to the silhouette branch of the network is a $240\times 320\times 5$ tensor, computed over }
%\ale{The silhouettes were first resized by a factor of 2 to speed up the computation. The average silhouettes were then computed over 
$5$ time intervals $\Delta t$, defined by,
\begin{equation}
\Delta t_k = \frac{T}{3^k},\text{ with } k=\left[0,\ ...,\ N\right] ,
\end{equation}
where $N=4$, and $T$ is the maximum buffer size in the multi-scale silhouette image, set to 1000 frames. {This choice of value for $T$ is explored in Section~\ref{sec:results}.}
%\ale{The 5 average silhouettes where then combined in a 3D tensor as described in Section~\ref{sec:gen_sil}. The value of $T$ defines the buffer size, i.e. the time length of the maximum interval over which the average silhouettes are computed. In order to choose $T$ appropriately, 4 different values were tested for 
{Data augmentation was performed using a rotation range of $\theta = \pm5^\circ$ and a random shift of $t_x = t_y = \pm20\%$ range. The silhouettes branch of the network architecture, depicted in Figure~\ref{fig:architecture_combined}, is formed by two stacks of sequential convolution-activation-pooling layers, followed by a fully connected layer producing the EE. The activation function adopted was a rectified linear unit (\textit{ReLu}), the pooling size was 2 and the stride length for each layer was also 2. Optimal parameters were found by training each network for 1000 epochs and selecting the model with the minimum validation loss after at least 30 epochs of training.}

\noindent \textbf{\em Accelerometers:}
Using the network proposed by Zhu \etal \cite{Zhu2015c} as a baseline for the accelerometer branch of \fullnet{}, we adopted their architecture 
%to investigate any possible improvement achievable. Following \cite{Zhu2015c}, we implemented 
of a multi-channel CNN that processes each component of the accelerometer independently, with two stacks of convolution-activation-pooling, using respectively 8 and 4 filters, with a kernel size of 5 and a stride length of 2. We replaced the \textit{tanh} activation function with a \textit{ReLu}, increased the input vector from 256 to 1000 elements and used both the wrist and waist mounted accelerometers as input, combining them into a single 6-channel input. {This produced a tensor input of size $1000\times 6$, which was fed into the accelerometer branch of the network}. In addition to that, we also estimated the gravity vector using a Wiener filter \cite{Rizun2008} 
%\DimaN{[any reference? As a vision reader, I don't know what this is]} 
with a window size of 1 second, and {subtracted its direction} from the accelerometer data. The baseline model Zhu \etal \cite{Zhu2015c} was implemented without the anthropometric feature vector (as we have no heart rate data available), and using both accelerometers as per \accnet{}. We show that each of these modifications allowed a better estimation of the EE in our tests. Our implementation of Zhu \etal \cite{Zhu2015c} has higher root mean square error (RMSE) than our proposed modified version in \accnet{} for 10 out of the 11 actions (excluding Wipe), as well as the overall error.

\section{Results}
\label{sec:results}
The proposed network \fullnet{} was tested using leave-one-subject-out cross-validation. {As baselines}, we also show the results obtained from (a) METs lookup tables \cite{AINSWORTH1993}, (b) {previous state-of-the-art on the same dataset from} Tao \etal \cite{Tao2018} which combined {hand-crafted} visual (full RGB-D images) and accelerometer {features with an SVM classifier}, and (c) the accelerometer network proposed by Zhu \etal \cite{Zhu2015c}. We also report results on single modalities: \accnet{} and \silnet{}. 
{Comparative results are presented in} Figure~\ref{fig:results_bar}, showing
%\mm{obtained by averaging the RMSE of all the subjects per activity} 
{the per-activity RMSE between the calories estimated (per minute) and the ground truth, obtained by averaging the errors for each activity class first, and then considering the mean across the subjects. The overall error was instead evaluated by averaging all the RMSEs regardless of the activity performed, by considering the mean across all the subjects}.
%\sout{We also present the per-activity average RMSE between the calories estimated (per minute) and the ground truth. These were computed by averaging the RMSE of all the subjects for each single activity class.} 

The figure shows that the EE estimation of the lookup table (METs) produces the highest error, with an overall RMSE of {1.50} cal/min when compared to the Calorimeter device. As already stated, METs tables are based on statistical measurements and are not suitable for subject-specific estimations. Tao \etal \cite{Tao2018}'s method improves over the the METs table, providing an overall average error of  1.30 cal/min. %However, the application of a deep-learning algorithm, as proposed by 
Zhu \etal \cite{Zhu2015c}, allows an overall improvement of the error {for most of the classes}, using accelerometer data only. %, showing that the high complexity of the EE estimation can be better tackled with deep learning techniques rather than with more classical computer vision methodologies.
When compared with the rest of the methods, our proposed \fullnet{} achieves the best results, producing an error which is almost {30\% lower than} the result from Zhu \etal \cite{Zhu2015c}, with a reduction of the RMSE from {1.21 to 0.88} cal/min. 

It order to stress the importance of our results, we also provide a comparison of our proposed method when accelerometers (\accnet{}) or silhouettes (\silnet{}) are used independently. Results for \accnet{} already show an overall reduction of the error from {1.21 to 0.98} cal/min {showing the advantage of our proposed modifications.} 
%when compared to Zhu \etal \cite{Zhu2015c}, which demonstrates that their architecture was not fine tuned (as stated by authors themselves \cite{Zhu2015c}). 
The error reduction is particularly pronounced for low-activity classes like \textit{Stand} and \textit{Sit}, which we believe to be due to the high pass gravity filter that we apply to the raw accelerometer signals. A further reduction of the error is achieved by \silnet{}, when silhouettes only are used for the EE, with an overall error of {0.95} cal/min. The RMSE of \silnet{} is particularly improved compared to \accnet{} especially for the \textit{Exercise} and \textit{Stretch} activity classes, {as these activities are} %probably due to their very strong visual signature which is 
better characterized by the video sensor.

{During our experiments, we noticed that all the methodologies tested struggled to estimate the calorie expenditure during the activities \textit{Exercise} and \textit{Stretch}. We believe this increased error is due to the high inter- and intra-class variance of these activities, estimated to be respectively 7.3 and 2.3 calories/min for the \textit{Exercise} class, and 4.0 and 1.0 calories/min for \textit{Stretch}. 
%As also depicted graphically in Figure~\ref{fig:dataset} \DimaN{[where is that in figure 4? The figures don't show where exercise or stretch activities are?]}, 
These values appear to be between 20 and 60 times higher than the variance shown by other classes like \textit{Sitting} or \textit{Walking}, as a consequence of the rather small training dataset available. A richer dataset including subjects with more different metabolisms and performing a wider range of activities would benefit the reduction of this error.} 

% These results are in agreement with the work in Tao \etal \cite{Tao2017}, which already demonstrated the superiority of a video-accelerometer fusion even in a traditional, non-deep learning approach. \mmn{Not sure if the latter point on fusion of the data undermines the novelty of this work here or not? Similarly above, when we say Zhu shows the importance of a deep learning approach.}

\begin{figure}
	% C:\Users\am14795\Dropbox\documenti\SPHERE\keras_play\calories\plot_final_results.py
	\centering
		\includegraphics[width=\textwidth]{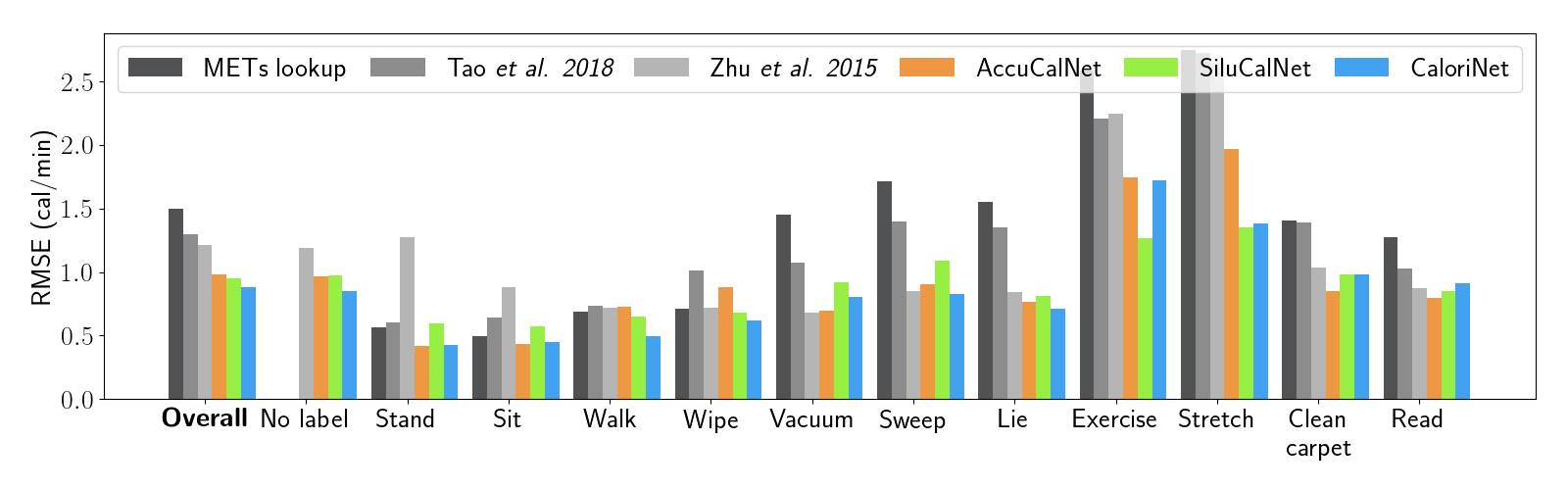}
	\caption{Results in terms of average per-activity RMSE for the calorie estimation.}
	\label{fig:results_bar}
\end{figure}

%\mmn{I think this section needs a remark on why the results here are novel, in terms of accuracy, use on non-RGB data, and the impact it can have on healthcare environments by returning back to the idea of privacy - to remind the reader why this is really important}. \ale{\textit{Added below.}}

{Sample qualitative results are} presented in Figure~\ref{fig:prediction_case}, which shows the continuous calorie prediction for a single individual, evaluated with different algorithms and compared with the ground truth. We observe very good agreement for \fullnet{} and \silnet{} with the ground truth, while Zhu \etal \cite{Zhu2015c}'s method shows quite erratic behaviour, missing the peak measurement of calories during the \textit{Exercise} 
%\DimaN{[either use capital letter or small letter for the activity throughout]}
activity (the red interval in the ground truth). The METs table only provides a step-wise prediction, as it only takes into account the labels of the activities performed, with data missing in those segments where no label was available.

\begin{figure}
	% C:\Users\am14795\Dropbox\documenti\SPHERE\keras_play\calories\plot_prediction_case.py
	\begin{center}
		\includegraphics[width=.88\textwidth]{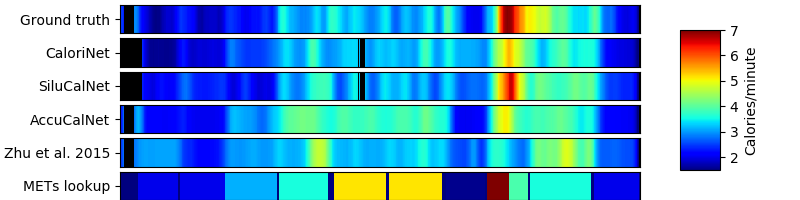}
	\end{center}
	\caption{Comparison of the calories measured for a single subject {(Subject 2, Session 2)} and the prediction obtained with different methods. Black lines depict missing data.}
	\label{fig:prediction_case}
\end{figure}

{%We assess the sensitivity of the method to the choice of the} buffer size $T$ presented in Section~\ref{sec:imp_det}. 
We evaluated the sensitivity of \fullnet{} when the buffer size parameter $T$ is varied. For this test, we adjusted $T$ to 250, 500, 1000 and 2000 frames, and evaluated the overall error for each buffer size. Results are presented in Figure~\ref{fig:error_vs_buffersize}, showing that {lower $T$ values produce inferior results while the method is performing consistently for $1000 \le T \le 2000$ frames.}% the best predictions are obtained for a value of $T\ge 1000$, indicating a consistent behaviour across all the activities. Since values greater than 1000 do not produce any noticeable improvement, but increase the risk of overfitting, we chose $T=1000$ frames for the optimal buffer size.}

%These results show the ability of our proposed \fullnet{} approach {as a robust tool in accurately estimating the EE within} private environments for use in healthcare and AAL applications. 
%As already mentioned earlier in this work, silhouettes and accelerometers constitute the only form of data available in privacy sensitive settings \cite{Woznowski2015}, with data being collected in this form from hundreds of houses \cite{Hall2016}. 
%Our work will enable the possibility of accurately estimating the EE from such houses while protecting the privacy of their households.

\begin{figure} 
	% C:\Users\am14795\Dropbox\documenti\SPHERE\keras_play\calories\plot_error_vs_buffersize.py
	\begin{center}
		\includegraphics[width=0.5\textwidth]{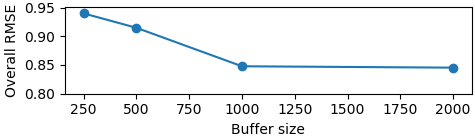}
	\end{center}
	\caption{Overall error of \fullnet{} for different buffer sizes.}
\label{fig:error_vs_buffersize}
\end{figure}

%\DimaN{[the buffer discussion should move here, and maybe should be studied relative to individual actions? Is there a different optimal T for the various actions?]} \ale{\textit{I had a look at the results per individual action, and they are consistent with the overall behaviour, with differences just due to random noise. Note added}}

%\DimaN{[Missing discussion on failure cases. What else needs to be improved? Why is the error in exercise still significantly higher?]} \ale{\textit{Because of the very high inter-class variation of the calories for this activity. I added a note for this}}

%\DimaN{[I have a concern in the way Figure 6 is reported. It has an implicit harmful effect that we figure out the activity first then we provide the results, which is not the case. This should be picked up throughout. What about frames for which there's no activity, how well do we do for these?]} \ale{\textit{we do not do any activity prediction, it's simply the way we show results. Figure 6 shows the continuous results for our method, maybe we should show this figure and discuss it before the per-activity results? Done}} 

\section{Conclusions}
The increasing adoption of healthcare monitoring devices in AAL environments demands the necessity of privacy-aware video systems. Here, we presented a novel, fused deep architecture for {online} estimation of energy expenditure using a combination of image silhouettes and accelerometer data. Systems recording such data are, for example, currently being deployed in one hundred homes \cite{Sphere100}. Silhouettes were first combined into a multi-channel average image, which provides temporal information for different time lengths. We then fed average silhouettes with accelerometer data in a CNN, that extracted features which were in turn fed into a fully connected layer that estimated the calories expended. {We obtained state-of-the-art results in comparison to other existing approaches while protecting privacy}. 
%Based on our results, we believe that our method can provide a perfect tool for monitoring the well-being of people in AAL houses, while protecting their privacy.
\bibliographystyle{plain}
\bibliography{ComputerVision-CaloriesExpenditure}

\begin{thebibliography}{10}

\bibitem{AINSWORTH1993}
Barbara~E. Ainsworth, William~L. Haskell, Arthur~S. Leon, David~R. Jacobs,
  Henry~J. Montoye, James~F. Sallis, and Ralph~S. Paffenbarger.
\newblock {Compendium of Physical Activities: classification of energy costs of
  human physical activities}.
\newblock {\em Medicine {\&} Science in Sports {\&} Exercise}, 25(1):71--80,
  1993.

\bibitem{Albinali2010}
Fahd Albinali, Stephen Intille, William Haskell, and Mary Rosenberger.
\newblock {Using wearable activity type detection to improve physical activity
  energy expenditure estimation}.
\newblock {\em Proceedings of the 12\textsuperscript{th} ACM international
  conference on Ubiquitous computing}, page 311, 2010.

\bibitem{Altini2012a}
Marco Altini, Julien Penders, and Oliver Amft.
\newblock {Energy expenditure estimation using wearable sensors}.
\newblock In {\em Proceedings of the conference on Wireless Health}, pages
  1--8, New York, New York, USA, 2012. ACM Press.

\bibitem{Altini2015a}
Marco Altini, Julien Penders, Ruud Vullers, and Oliver Amft.
\newblock {Estimating Energy Expenditure Using Body-Worn Accelerometers: A
  Comparison of Methods, Sensors Number and Positioning}.
\newblock {\em IEEE Journal of Biomedical and Health Informatics},
  19(1):219--226, January 2015.

\bibitem{Armstrong1990}
Neil Armstrong, John Balding, Peter Gentle, Joanne Williams, and Brian Kirby.
\newblock {Peak Oxygen Uptake and Physical Activity in I I to 16-Year-Olds}.
\newblock {\em Pediatric Exercise Science}, 2(20):349--358, 1990.

\bibitem{Bilen2016}
Hakan Bilen, Basura Fernando, Efstratios Gavves, Andrea Vedaldi, and Stephen
  Gould.
\newblock {Dynamic Image Networks for Action Recognition}.
\newblock {\em IEEE Conference on Computer Vision and Pattern Recognition},
  pages 3034--3042, 2016.

\bibitem{Birchley2017}
Giles Birchley, Richard Huxtable, Madeleine Murtagh, Ruud {Ter Meulen}, Peter
  Flach, and Rachael Gooberman-Hill.
\newblock {Smart homes, private homes? An empirical study of technology
  researchers' perceptions of ethical issues in developing smart-home health
  technologies}.
\newblock {\em BMC Medical Ethics}, 18(1):1--13, 2017.

\bibitem{Bobick2001}
A.~F. Bobick and J.~W. Davis.
\newblock {The recognition of human movement using temporal templates}.
\newblock {\em IEEE Transactions on Pattern Analysis and Machine Intelligence},
  23(3):257--267, March 2001.

\bibitem{Byrne1993}
A.~Byrne and D.~G. Byrne.
\newblock {The effect of exercise on depression, anxiety and other mood states:
  A review}.
\newblock {\em Journal of Psychosomatic Research}, 37(6):565--574, 1993.

\bibitem{Cao2016}
Zhe Cao, Tomas Simon, Shih-En Wei, and Yaser Sheikh.
\newblock {Realtime Multi-Person 2D Pose Estimation using Part Affinity
  Fields}.
\newblock 2016.

\bibitem{Ceesay1989}
Sana~M. Ceesay, Andrew~M. Prentice, Kenneth~C. Day, Peter~R. Murgatroyd,
  Gail~R. Goldberg, Wendy Scott, and G.~B. Spurr.
\newblock {The use of heart rate monitoring in the estimation of energy
  expenditure: a validation study using indirect whole-body calorimetry}.
\newblock {\em British Journal of Nutrition}, 61(02):175, March 1989.

\bibitem{Celi2010}
Francesco~S Celi, Robert~J Brychta, Joyce~D Linderman, Peter~W Butler, A.~T.
  Alberobello, Sheila Smith, Amber~B Courville, Edwin~W Lai, Rene Costello,
  M.~C. Skarulis, G.~Csako, A.~Remaley, K.~Pacak, and K.~Y. Chen.
\newblock {Minimal changes in environmental temperature result in a significant
  increase in energy expenditure and changes in the hormonal homeostasis in
  healthy adults}.
\newblock {\em European Journal of Endocrinology}, 163(6):863--872, December
  2010.

\bibitem{Chodzko-Zajko2009}
Wojtek~J. Chodzko-Zajko, David~N. Proctor, Maria~A. {Fiatarone Singh},
  Christopher~T. Minson, Claudio~R. Nigg, George~J. Salem, and James~S.
  Skinner.
\newblock {Exercise and physical activity for older adults}.
\newblock {\em Medicine and Science in Sports and Exercise}, 41(7):1510--1530,
  2009.

\bibitem{Craig2003}
Cora~L. Craig, Alison~L. Marshall, Michael Sj{\"{o}}str{\"{o}}m, Adrian~E.
  Bauman, Michael~L. Booth, Barbara~E. Ainsworth, Michael Pratt, Ulf Ekelund,
  Agneta Yngve, James~F. Sallis, and Pekka Oja.
\newblock {International physical activity questionnaire: 12-Country
  reliability and validity}.
\newblock {\em Medicine and Science in Sports and Exercise}, 35(8):1381--1395,
  2003.

\bibitem{Ellis2014}
Katherine Ellis, Jacqueline Kerr, Suneeta Godbole, Gert Lanckriet, David Wing,
  and Simon Marshall.
\newblock {A random forest classifier for the prediction of energy expenditure
  and type of physical activity from wrist and hip accelerometers}.
\newblock {\em Physiological Measurement}, 35(11):2191--2203, December 2014.

\bibitem{Gibson2018}
Eli Gibson, Wenqi Li, Carole Sudre, Lucas Fidon, Dzhoshkun~I. Shakir, Guotai
  Wang, Zach Eaton-Rosen, Robert Gray, Tom Doel, Yipeng Hu, Tom Whyntie,
  Parashkev Nachev, Marc Modat, Dean~C. Barratt, S{\'{e}}bastien Ourselin,
  M.~Jorge Cardoso, and Tom Vercauteren.
\newblock {NiftyNet: a deep-learning platform for medical imaging}.
\newblock {\em Computer Methods and Programs in Biomedicine}, 158:113--122, May
  2018.

\bibitem{Hall2016}
J.~Hall, S.~Hannuna, M.~Camplani, M.~Mirmehdi, D.~Damen, T.~Burghardt, L.~Tao,
  A.~Paiement, and I.~Craddock.
\newblock {Designing a video monitoring system for AAL applications: The SPHERE
  case study}.
\newblock In {\em IET Conference Publications}, volume 2016, pages 126--126.
  Institution of Engineering and Technology, 2016.

\bibitem{He2014}
Kaiming He, Xiangyu Zhang, Shaoqing Ren, and Jian Sun.
\newblock {Spatial Pyramid Pooling in Deep Convolutional Networks for Visual
  Recognition}.
\newblock {\em IEEE Transactions on Pattern Analysis and Machine Intelligence},
  37(9):1904--1916, September 2015.

\bibitem{Jancke2001}
Gavin Jancke, Gina~D. Venolia, Jonathan Grudin, Jonathan~J. Cadiz, and Anoop
  Gupta.
\newblock {Linking Public Spaces - Technical and Social Issues}.
\newblock {\em Proceedings of the International Conference on Human Factors in
  Computing Systems}, (3):530--537, 2001.

\bibitem{Kong2012}
Fanyu Kong and Jindong Tan.
\newblock {DietCam: Automatic dietary assessment with mobile camera phones}.
\newblock {\em Pervasive and Mobile Computing}, 8(1):147--163, 2012.

\bibitem{Mahendran2017}
Siddharth Mahendran, Haider Ali, and Rene Vidal.
\newblock {3D Pose Regression Using Convolutional Neural Networks}.
\newblock In {\em IEEE Conference on Computer Vision and Pattern Recognition
  Workshops}, pages 494--495. IEEE, July 2017.

\bibitem{Marshall2011}
Simon~J. Marshall and Ernesto Ramirez.
\newblock {Reducing Sedentary Behavior: A New Paradigm in Physical Activity
  Promotion}.
\newblock {\em American Journal of Lifestyle Medicine}, 5(6):518--530, 2011.

\bibitem{Massa2016}
Francisco Massa, Renaud Marlet, and Mathieu Aubry.
\newblock {Crafting a multi-task CNN for viewpoint estimation}.
\newblock pages 1--12, September 2016.

\bibitem{Myers2015}
Austin Myers, Nick Johnston, Vivek Rathod, Anoop Korattikara, Alex Gorban,
  Nathan Silberman, Sergio Guadarrama, George Papandreou, Jonathan Huang, and
  Kevin Murphy.
\newblock {Im2Calories: Towards an Automated Mobile Vision Food Diary}.
\newblock In {\em IEEE International Conference on Computer Vision}, volume
  2015 Inter, pages 1233--1241. IEEE, December 2015.

\bibitem{Niu2016}
Zhenxing Niu, Mo~Zhou, Le~Wang, Xinbo Gao, and Gang Hua.
\newblock {Ordinal Regression with Multiple Output CNN for Age Estimation}.
\newblock In {\em IEEE Conference on Computer Vision and Pattern Recognition},
  pages 4920--4928. IEEE, June 2016.

\bibitem{Pande2015}
Amit Pande, Jindan Zhu, Aveek~K. Das, Yunze Zeng, Prasant Mohapatra, and Jay~J.
  Han.
\newblock {Using Smartphone Sensors for Improving Energy Expenditure
  Estimation}.
\newblock {\em IEEE Journal of Translational Engineering in Health and
  Medicine}, 3(September), 2015.

\bibitem{Sphere100}
SPHERE Project.
\newblock {SPHERE 100 Homes Study}.
\newblock \url{http://irc-sphere.ac.uk/100-homes-study}, 2018.

\bibitem{Ray2014}
P.~P. Ray.
\newblock {Internet of Things based Physical Activity Monitoring (PAMIoT): An
  Architectural Framework to Monitor Human Physical Activity}.
\newblock {\em IEEE Calcutta Conference}, pages 32--34, 2014.

\bibitem{Rizun2008}
Peter Rizun.
\newblock {Optimal Wiener Filter for a Body Mounted Inertial Attitude Sensor}.
\newblock {\em Journal of Navigation}, 61(03):455--472, jul 2008.

\bibitem{Staudenmayer2009a}
J~Staudenmayer, D~Pober, S~E Crouter, D~R Bassett, and P~Freedson.
\newblock {An artificial neural network to estimate physical activity energy
  expenditure and identify physical activity type from an accelerometer}.
\newblock {\em Journal of Applied Physiology}, (17):1300--1307, 2009.

\bibitem{Tan2016}
Li~Kuo Tan, Yih~Miin Liew, Einly Lim, and Robert~A. McLaughlin.
\newblock {Cardiac left ventricle segmentation using convolutional neural
  network regression}.
\newblock In {\em IEEE Conference on Biomedical Engineering and Sciences},
  pages 490--493. IEEE, December 2016.

\bibitem{Tao2018}
Lili Tao, Tilo Burghardt, Majid Mirmehdi, Dima Damen, Ashley Cooper, Massimo
  Camplani, Sion Hannuna, Adeline Paiement, and Ian Craddock.
\newblock {Energy expenditure estimation using visual and inertial sensors}.
\newblock {\em IET Computer Vision}, 12(1):36--47, February 2018.

\bibitem{Tao2017}
Lili Tao, Tilo Burghardt, Majid Mirmehdi, Dima Damen, Ashley Cooper, Sion
  Hannuna, Massimo Camplani, Adeline Paiement, and Ian Craddock.
\newblock {Calorie Counter: RGB-Depth Visual Estimation of Energy Expenditure
  at Home}.
\newblock {\em Lecture Notes in Computer Science}, 10116 LNCS:239--251, July
  2016.

\bibitem{Tao2016}
Lili Tao, Adeline Paiement, Dima Damen, Majid Mirmehdi, Sion Hannuna, Massimo
  Camplani, Tilo Burghardt, and Ian Craddock.
\newblock {A comparative study of pose representation and dynamics modelling
  for online motion quality assessment}.
\newblock {\em Computer Vision and Image Understanding}, 148:136--152, July
  2016.

\bibitem{Um2017}
Terry~Taewoong Um, Franz Michael~Josef Pfister, Daniel Pichler, Satoshi Endo,
  Muriel Lang, Sandra Hirche, Urban Fietzek, and Dana Kuli{\'{c}}.
\newblock {Data Augmentation of Wearable Sensor Data for Parkinson's Disease
  Monitoring using Convolutional Neural Networks}.
\newblock 2017.

\bibitem{Wang2018}
Baodong Wang, Lili Tao, Tilo Burghardt, and Majid Mirmehdi.
\newblock {Calorific Expenditure Estimation Using Deep Convolutional Network
  Features}.
\newblock In {\em 2018 IEEE Winter Applications of Computer Vision Workshops
  (WACVW)}, pages 69--76. IEEE, mar 2018.

\bibitem{Woznowski2015}
Przemyslaw Woznowski, Xenofon Fafoutis, Terence Song, Sion Hannuna, Massimo
  Camplani, Lili Tao, Adeline Paiement, Evangelos Mellios, Mo~Haghighi, Ni~Zhu,
  Geoffrey Hilton, Dima Damen, Tilo Burghardt, Majid Mirmehdi, Robert
  Piechocki, Dritan Kaleshi, and Ian Craddock.
\newblock {A multi-modal sensor infrastructure for healthcare in a residential
  environment}.
\newblock In {\em IEEE International Conference on Communication Workshop},
  pages 271--277. IEEE, June 2015.

\bibitem{Yang2010}
Che~Chang Yang and Yeh~Liang Hsu.
\newblock {A review of accelerometry-based wearable motion detectors for
  physical activity monitoring}.
\newblock {\em Sensors}, 10(8):7772--7788, 2010.

\bibitem{Zhu2015c}
Jindan Zhu, Amit Pande, Prasant Mohapatra, and Jay~J Han.
\newblock {Using Deep Learning for Energy Expenditure Estimation with Wearable
  Sensors}.
\newblock {\em 17\textsuperscript{th} International Conference on E-health
  Networking, Application {\&} Services}, pages 501--506, 2015.

\bibitem{Ziefle2011}
Martina Ziefle, Carsten R{\"{o}}cker, and Andreas Holzinger.
\newblock {Medical technology in smart homes: Exploring the user's perspective
  on privacy, intimacy and trust}.
\newblock {\em International Computer Software and Applications Conference},
  pages 410--415, 2011.

\end{thebibliography}

\end{document}